\documentclass{llncs}
\usepackage{makeidx}  
\usepackage[pdftex]{hyperref}
\usepackage{url}
\usepackage[pdftex]{graphicx,color}
\usepackage{bm}
\begin{document}
%
%
%
%
%
%
\mainmatter              
%
\title{Biologically Inspired \\ Feedforward Supervised Learning \\
for Deep Self-Organizing Map Networks}
%
%
\author{Takashi Shinozaki\inst{1,2}}
\authorrunning{Takashi Shinozaki} 
%
\tocauthor{Takashi Shinozaki}
\institute{CiNet, 
National Institute of Information and Communications Technology \\ 
1-4 Yamadaoka, Suita,
Osaka 565-0871, Japan, \\
\and
Graduate School of Information Science and Technology, Osaka University \\
1-5 Yamadaoka, Suita,
Osaka 565-0871, Japan, \\
\email{tshino@nict.go.jp}
}

\maketitle              

%
\begin{abstract}
In this study, we propose a novel deep neural network and 
its supervised learning method that uses a feedforward supervisory signal. 
The method is inspired by the human visual system 
and performs human-like association-based learning 
without any backward error propagation.
The feedforward supervisory signal that produces the correct result
is preceded by the target signal
and associates its confirmed label 
with the classification result of the target signal. 
It effectively uses a large amount of information from the feedforward
signal, and forms a continuous and rich learning representation.  
The method is validated 
using visual recognition tasks on the MNIST handwritten dataset.
\end{abstract}

\keywords{supervised learning, deep learning, self-organizing map}

\section{Introduction}
A multilayered deep neural network is one of the most powerful methods for 
human-like recognition tasks, such as
image \cite{Le2012} and speech recognition \cite{Dahl2012}. 
Some previous studies have demonstrated great performance
for supervised learning in signal classification tasks 
\cite{LeCun1989,Krizhevsky2012}. 
Gradient-based learning rules, in particular, 
back-propagation (BP) learning \cite{Rumelhart1986},
 are generally used for supervised learning in feedforward type networks.
However, the amount of supervisory information in the last layer is 
not sufficient to supervise the entire deep neural network 
because the information is selected and reduced from layer to layer. 
This tendency is more serious in pattern discrimination tasks 
because the amount of information is extremely limited 
to the discrete values of the discriminant label output. 
Bengio {\it et al.} proposed a stacked auto-encoder 
to ensure the amount of information from error signals 
by reconstructing the input and using layer-wise learning
\cite{Bengio2007}. 
However, layer-wise learning requires step-by-step learning, which
results in difficulties in incremental learning and online updating. 
Some previous studies have used unsupervised learning 
that does not use the prior information of the data structure, 
and reported self-organizing behavior and
good discrimination results in a very deep neural network
\cite{Le2012,Fukushima1980}. 
However, unsupervised learning could not control 
the classification of input data, which resulted in the 
enlargement of the network and low efficiency for learning. 

In this study, we propose a novel learning method for deep neural networks 
that uses 
feedforward propagated supervisory signals. 
The method effectively uses a large amount of information 
from the feedforward propagated supervisory signal, which
enables robust leaning in a deep neural network. 
It associates the classification of new input with 
that of pre-trained input, and revises the internal representation 
of the entire neural network.
We validate the propose learning method using a numerical simulation of 
visual pattern discrimination tasks.


\section{Network Model}
The network model was inspired by the human visual system in the cortex. 
The network is composed of self-organizing map (SOM) modules \cite{som}. 
Each SOM module consists of one hundred neurons, 
and receives a subset of the output of the corresponding location 
of the previous layer.
The connection is similar to a receptive field (RF) of recent convolutional 
neural networks, but is not convolutional, meaning no weight sharing 
among modules in a layer. 

Each neuron calculates an inner product between the weight and input 
as follows:
\begin{eqnarray}
\bm{u}_{l,t}=\bm{W}_{l,t} \bm{z}_{l-1,t},
\label{eq:u}
\end{eqnarray}
where $\bm{u}_{l,t}$ is the inner product of the $l$-th layer at time $t$, 
$\bm{W}_{l,t}$ is the weight matrix, and 
$\bm{z}_{l-1,t}$ is the output vector of the previous layer.
The inner product is then processed using 
winners-share-all (WSA) regularization in each module.
WSA is a variant of winner-takes-all (WTA), 
and involves not only the winning neuron, but also neighboring neurons. 
The neuron that has the most prominent inner product 
is selected as the winning neuron, and outputs 1.0. 
Neighboring neurons output a distance-decayed value 
determined using the Gaussian kernel. 
The output of the $l$-th layer is described as follows: 
\begin{equation}
  z_{l,t,j} = \exp(-d_j^2/2\sigma_{learn}^2),
\end{equation}
where $z_{l,t,j}$ is the $j$-th element of the output vector, 
and $d_j$ is the spatial distance from the winning neuron 
to the $j$-th neuron. 
The spatial decay factor $\sigma_{learn}$ 
is determined empirically, and set to $0.8$.

\begin{figure}[tb]
  \raisebox{0.20\linewidth}{\bf \large (a)}
  \includegraphics[width=0.55\linewidth]{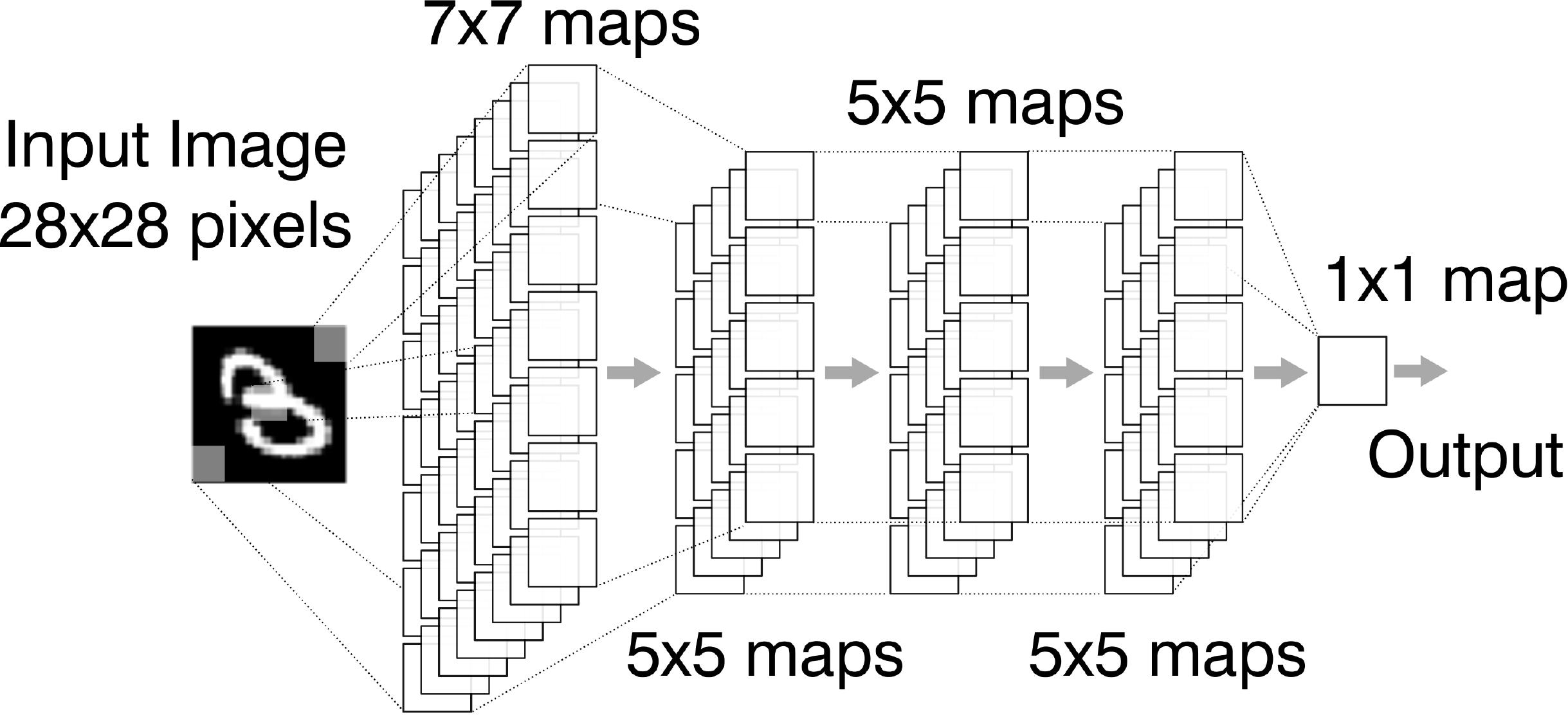}
  \hspace{0.2cm}
  \raisebox{0.20\linewidth}{\bf \large (b)} \hspace{0.1cm}
  \includegraphics[width=0.24\linewidth]{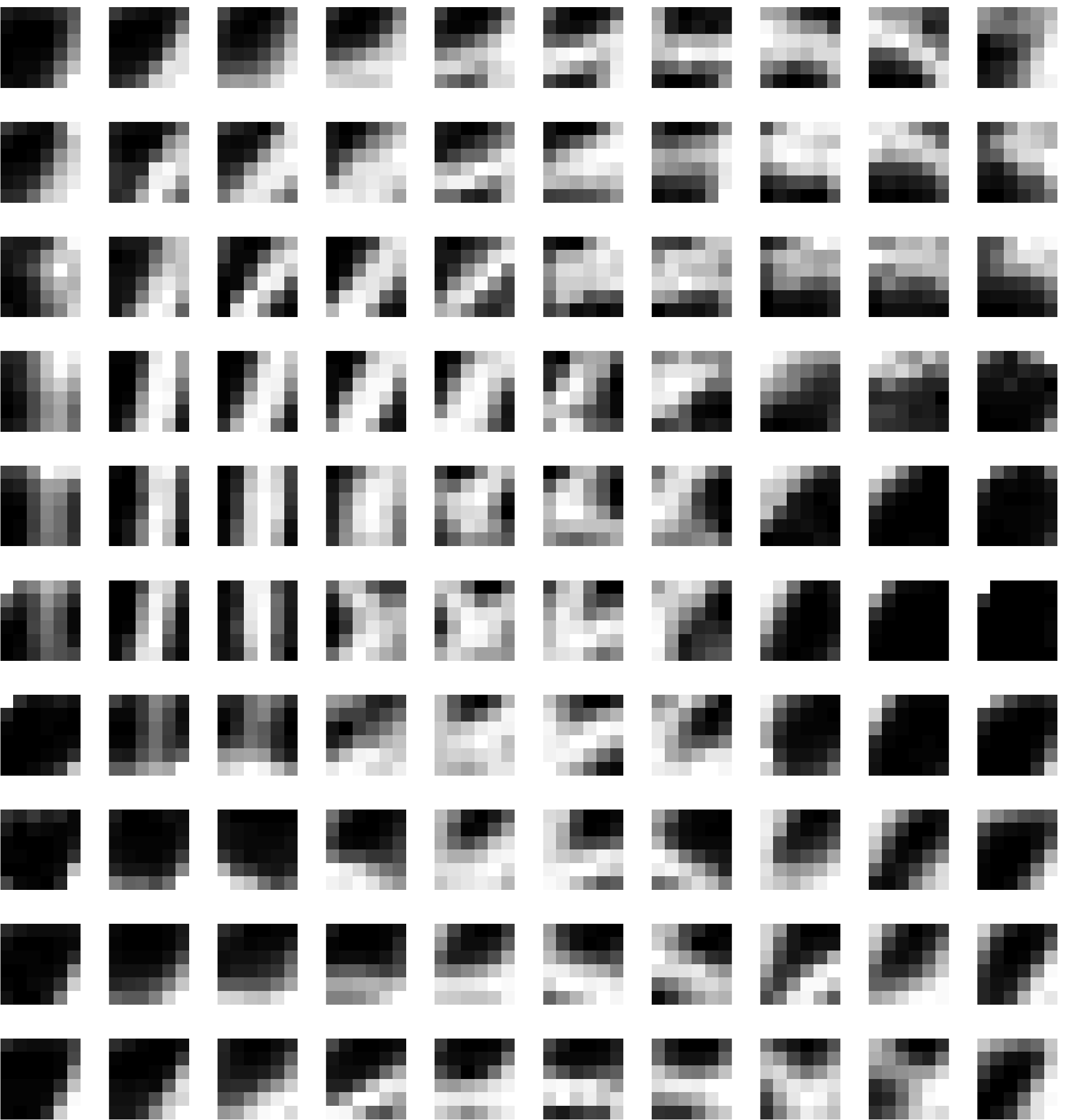}
  \hspace{-0.2cm}
  \caption{
    (a) Schematic of the network structure. 
    (b) Learning representation with continuous features in the first layer 
    generated by unsupervised pre-training. 
  }
  \label{fig:struct}
\end{figure}

Fig. \ref{fig:struct}(a) shows a schematic of the network structure 
for the experiment. 
As shown in Table \ref{tab:paras},
the first layer consists of 49 SOMs with 4,900 neurons, 
and neurons receive a 6$\times$6-pixel image that is a part of 
the input image of 28$\times$28 pixels, which
results in a total of 176,400 connections in the layer.

\begin{table}[tb]
  \begin{center}
    \begin{tabular}{ccrllr}
      Layer & Maps & \# of Neurons & RF & Stride & \# of Connections \\
      \hline
      1 & 7$\times$7 & 4,900 & 6$\times$6 pixels 
      & 4$\times$4 pixels & 176,400 \\
      2 & 5$\times$5 & 2,500 & 3$\times$3 maps & 1$\times$1 map & 2,250,000 \\
      3 & 5$\times$5 & 2,500 & 5$\times$5 maps & 1$\times$1 map & 6,250,000 \\
      4 & 5$\times$5 & 2,500 & 5$\times$5 maps & 1$\times$1 map & 6,250,000 \\
      5 & 1$\times$1 & 100 & 5$\times$5 maps & 1$\times$1 map & 250,000 \\
    \end{tabular}
  \end{center}
  \vspace{-0.2cm}
  \caption{Network parameters.}
  \label{tab:paras}
\end{table}

\section{Pre-training}
We use traditional unsupervised competitive learning for pre-training
\cite{som,compete}.
It updates the weight of the most prominent neuron and its neighbors, 
and forms a two-dimensional spatial structure of the template sets 
for the input pattern. The update rule is described as follows:
\begin{eqnarray}
\Delta \bm{w}_{l,t,j} = 
\rho_{pre} \exp(-d_j^2/2\sigma_{pre}^2) \bm{z}_{l-1,t},
\label{eq:compete}
\end{eqnarray}
where $\bm{w}_{l,t,j}$ is the weight vector, 
which is the $j$-th row of vector matrix $\bm{W}_{l,t}$, 
and $\bm{u}_{l,t}$ is the input vector to the module. 
As for traditional SOMs, the learning coefficient $\rho_{pre}$ linearly 
decreases from 1.00 at the beginning to 0.00 at the end, 
and the standard deviation of Gaussian kernel $\sigma_{pre}$ 
also decreases, from 3.5 to 0.0. 
The weight vector is normalized by L2-norm at every update. 
The method generates a spatially continuous feature map, 
which is similar to the map generated using
topographic independent component analysis 
\cite{tica}. 

Pre-training was performed in a layer-wise manner, 
which is similar to a biological critical period. 
Initially, only the first layer was learned, with 2,500 iterations. 
Next, 2,500 iterations were applied to the first and second layer. 
Finally, the first four layers were
processed, applying 10,000, 7,500, 5,000 and 2,500 iterations sequentially. 
The last layer was not processed by the pre-training.
Fig. \ref{fig:struct}(b) shows a typical example of the generated feature maps 
in the first layer. 

\begin{figure}[tb]
  \begin{center}
    \raisebox{0.20\linewidth}{\bf \large (a)}\hspace{-2.0ex}
    \includegraphics[height=0.23\linewidth]{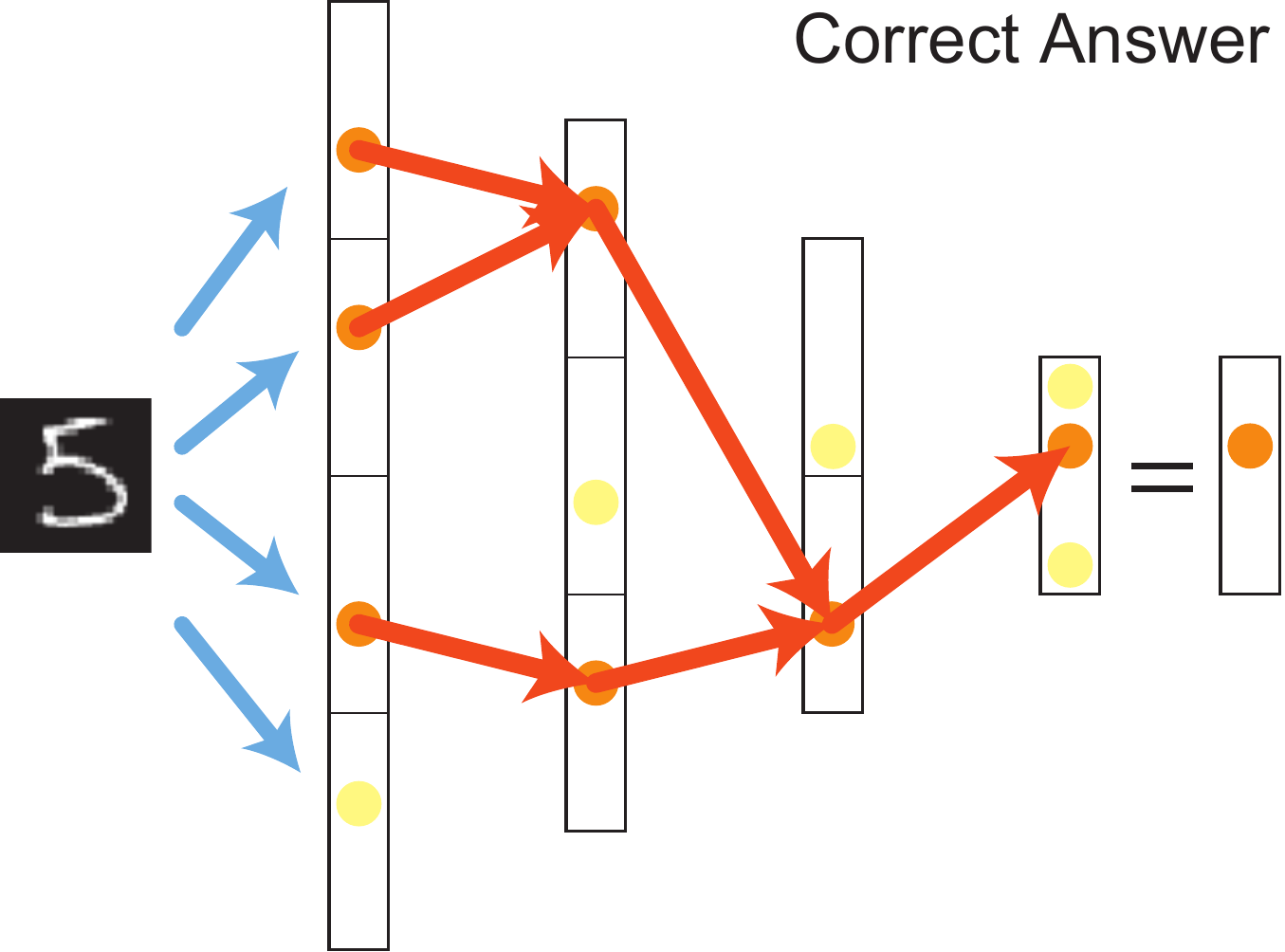}
    \hspace{0.06\linewidth}
    \raisebox{0.20\linewidth}{\bf \large (b)}\hspace{-2.0ex}
    \includegraphics[height=0.23\linewidth]{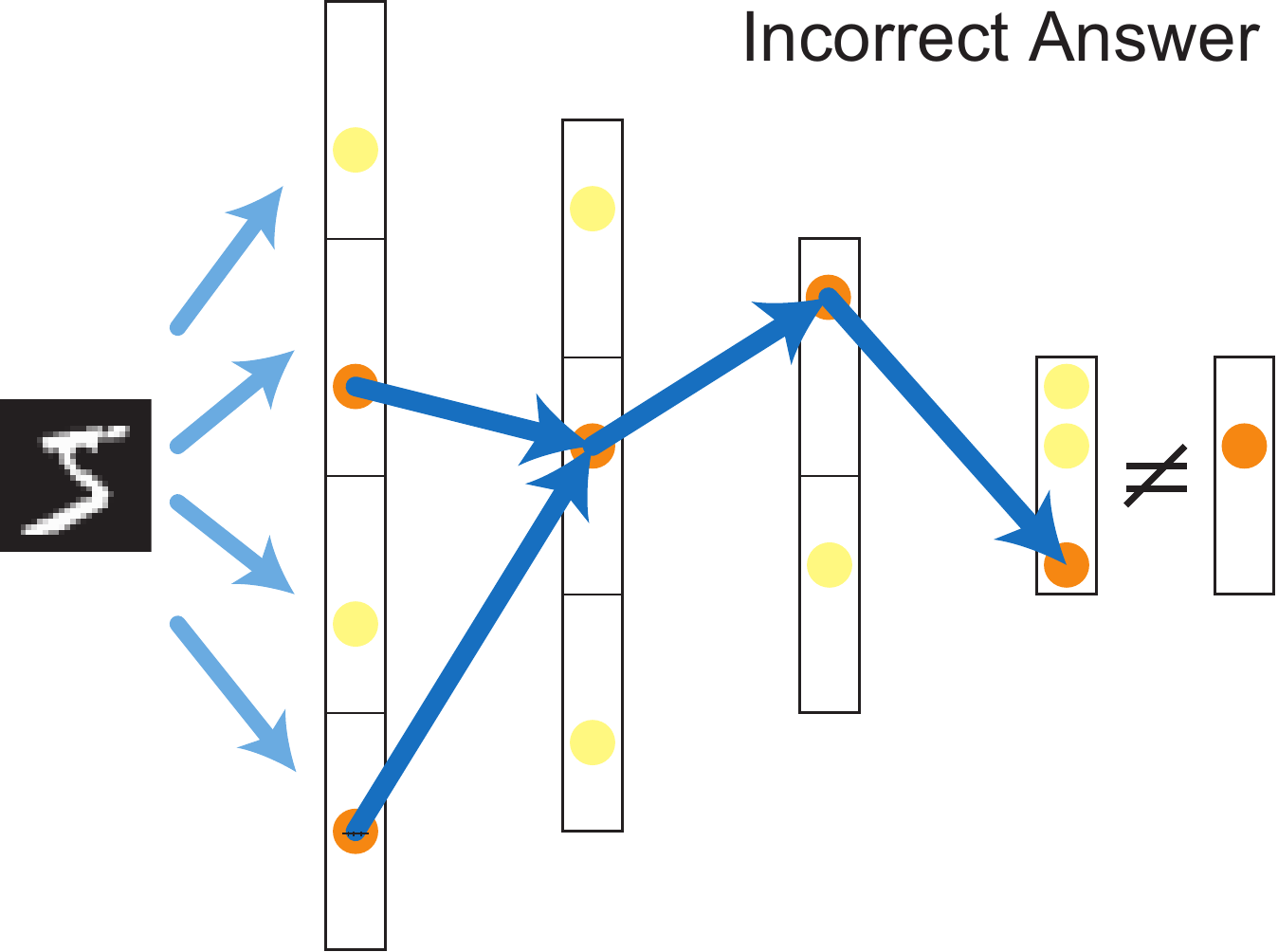}
    \vspace{0.5cm}\\
    \raisebox{0.20\linewidth}{\bf \large (c)}\hspace{-2.0ex}
    \includegraphics[height=0.23\linewidth]{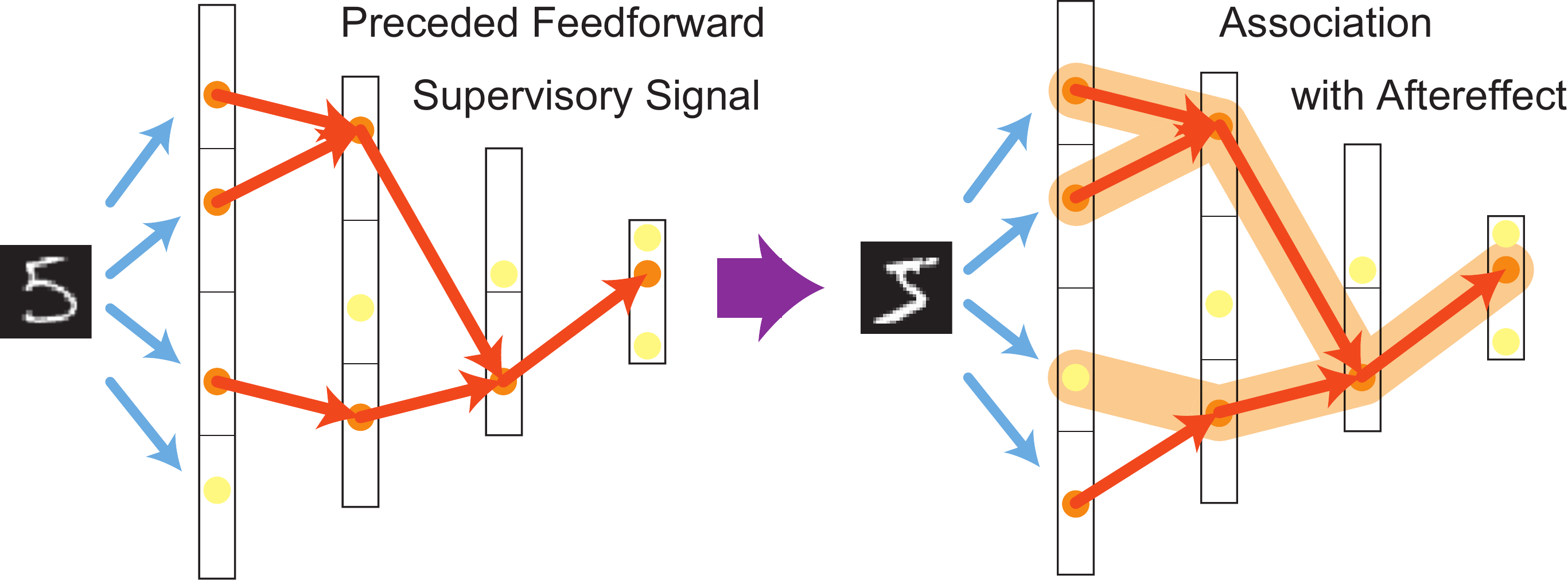}
  \end{center}
  \caption{
    (a) Example of a correct label output by a clearer image 
    that conducted a correct propagation `path'. 
    (b) Example of an incorrect label output by a difficult image. 
    (c) The difficult image produced the required label output
    with a guide of the after-effect conducted by the clearer image. 
    The shaded region retained the after-effect along the correct `path' 
    conducted by advance propagation.
  }
  \label{fig:learn}
\end{figure}

\section{Advance Propagation Learning}
Advance propagation (AP) learning is a supervised learning method 
that enables a feedforward supervisory signal using 
the sparse dynamics of the network. 
It is based on learning vector quantization (LVQ) \cite{lvq}, 
but requires additional advance input as a supervisory signal 
to specify the learning `location'.
Before processing the target input, advance input, 
which produces the required classification label, 
is propagated throughout the entire network. 
Then, the target input is processed 
with the after-effect of the advance input. 
The after-effect guides the correct `path' of the propagation, 
and specifies the learning `location' in the network. 
The point is that advance propagation 
does not restrict the propagation 'path', 
but just suggests it. It merges various paths by various types of inputs 
with a same internal learning representation. 
LVQ-like conditional learning followed by the target input
specifies the learning `direction',
thereby revising the weight vector to produce the required label. 

A learning trial is processed as follows: 
First, the target input $\bm{z}_{l-1,t}$ is processed 
by each module in the network, 
and the output label of the network is checked. 
If it corresponds to the required label, then the
weights of the activated units 
are updated by competitive learning as Eq.\ref{eq:compete} 
(Fig. \ref{fig:learn}(a)).
Otherwise, the weights are updated 
to opposite direction of Eq.\ref{eq:compete}
(Fig. \ref{fig:learn}(b)), and then AP learning is evoked. 
First, the advance input that produces the required label output 
is processed by the same network, which
results in the required label output at time $t-1$ 
(Fig. \ref{fig:learn}(c)left). 
Subsequently, the target input is processed again 
with the after-effect of the advance input
(Fig. \ref{fig:learn}(c)right) as follows: 
\begin{eqnarray}
\bm{z'}_{l,t} = \beta\bm{z}_{l-1,t-1} + (1-\beta)\bm{z}_{l-1,t},
\label{eq:input}
\end{eqnarray}
where $\beta$ is the ratio of the after-effect of the advance input. 
The vector represents the direction of the feature vector $\bm{z}_{l-1,t}$
corrected by the after-effect of the advance input $\bm{z}_{l-1,t-1}$. 
The important point is that the network has highly nonlinear behavior 
using WSA, and the output is not equal to that produced by 
the linear summation of the two inputs. 
The following competitive learning uses the combined input in
as same manner as Eq. \ref{eq:compete}. 
Consequently, 
the full version of the equation with 
multi-layer decay and a Gaussian kernel for the WSA output is 
described as follows: 
\begin{eqnarray}
\Delta \bm{w}_{l,t,j} = 
r^{n-l} \rho_{base} \exp(-d_j^2/2\sigma_{learn}^2) 
\{\beta \bm{z}_{l-1,t-1}+(1-\beta) \bm{z}_{l-1,t}\},
\label{eq:apl}
\end{eqnarray}
where $r$ is the decay coefficient from layer to layer
and $n$ is the total number of layers. 
The weight vector is normalized by L2-norm at every update, 
as in traditional competitive learning. 
We used $(\rho_{base},\beta,r,n,\sigma_{learn})=(0.20,0.4,0.7,5,0.4)$ 
in the experiments. 

\section{Experiments}
To validate the proposed learning method, 
we performed a discrimination test on the MNIST handwritten image dataset
(10 digits, 28$\times$28 pixels, grayscale) \cite{Mnist}.
AP learning was applied to the pre-trained network.  
The most matched output of the last layer in the pre-training result 
was selected for each label. 
Advance inputs as supervisory signals were 
dynamically determined, and updated from one trial to the next. 
Each input signal was initially tested using its label, 
and AP learning was applied if the label was incorrect.  

One learning block consisted of 10,000 samples of the 
training dataset input for learning, and 
10,000 samples of the validation dataset 
only to calculate the error rate. 
The calculation was performed on a workstation (Opteron 6366 1.8 GHz)
using custom C code with OpenMP parallelization.

\begin{figure}[tb]
  \raisebox{0.27\linewidth}{\bf \large (a)}
  \hspace{-0.2cm}
  \includegraphics[width=0.4\linewidth]{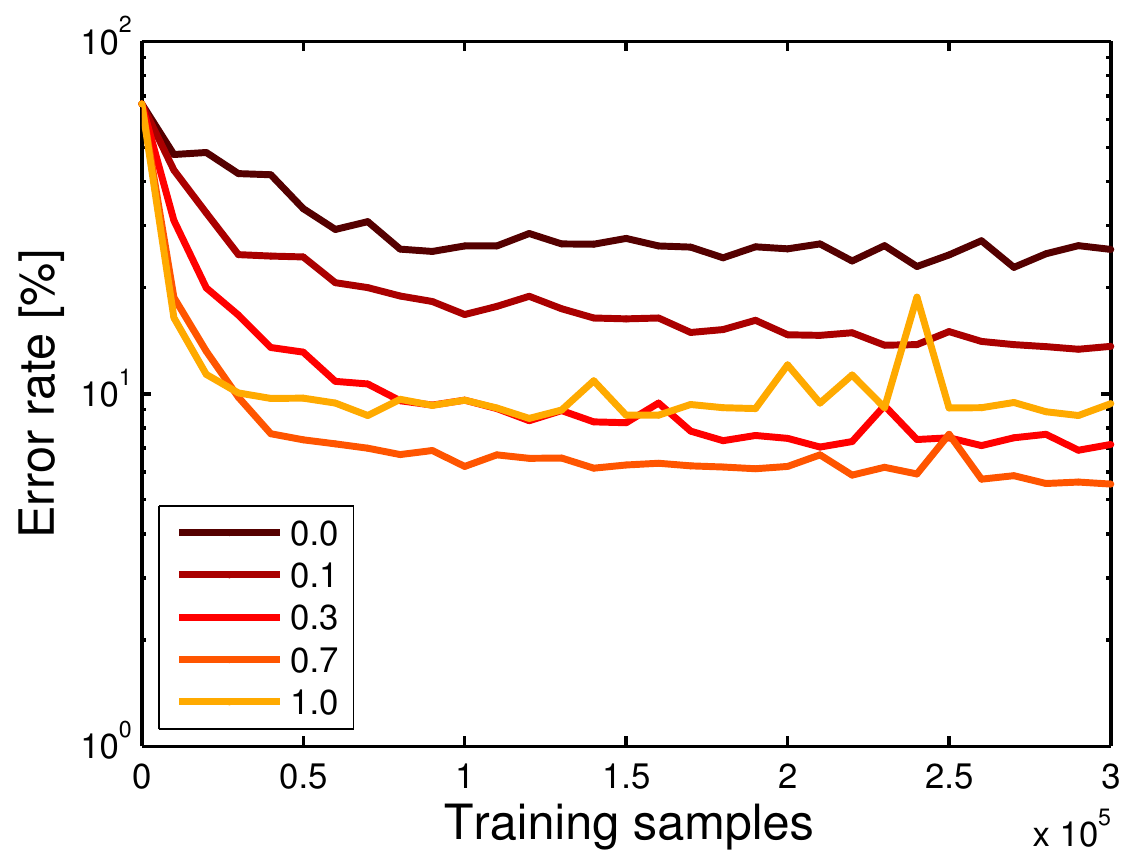}
  \hspace{0.1cm}
  \raisebox{0.27\linewidth}{\bf \large (b)}
  \hspace{-0.6cm}
  \raisebox{0.03\linewidth}{
  \includegraphics[width=0.5\linewidth]{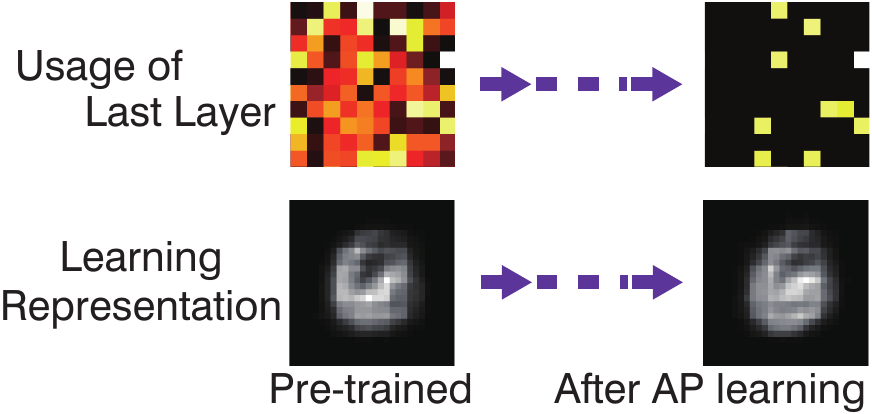}}

  \caption{
    (a) Error rate of the discrimination for each layer decay parameter $r$. 
    (b) {\it upper}: Use of neurons in the last layer. 
    Brighter color represents the use of the corresponding neuron. 
    Ten representative neurons corresponding respective digit labels
    remained after AP learning. 
     {\it lower}: Optimal stimuli of a typical representative neuron 
    (which codes the figure `6') before and after AP learning. 
  }
  \label{fig:misc}
\end{figure}

Fig. \ref{fig:misc}(a) represents 
the change in the error rate using AP learning. 
Initially, the error rate was determined using 
pre-training with unsupervised competitive learning, which
resulted in $66.5$ \%. 
AP learning improved the error rate
to $3.8$ \% after 20 iterations of the entire training set 
with the decay coefficient $r=0.7$.
If the decay coefficient equaled zero, which meant that there was no learning 
in the upper layer, then the learning stopped at quite an early stage, which
resulted in a high error rate ($22.5$ \%). 
If the coefficient value was non-zero, then learning was
processed over the entire network, which resulted in a low error rate. 
The results demonstrated that the proposed method 
effectively processed learning over the entire network at the time.

The initial scattered learning representation in the last layer
was rearranged to a more sparse and efficient style 
(Fig. \ref{fig:misc}(b){\it upper}).
Simultaneously, the optimal stimuli of the representative neurons 
were modified to create a more generalized image 
(Fig. \ref{fig:misc}(b){\it lower}).

\section{Conclusion}
In this paper, we introduced a novel supervised learning method
for a deep feedforward neural network, and validated the efficiency 
for a visual recognition task. 
The method focused on using the rich input information 
in the early layer as the supervisory signal in each layer.  
We demonstrated that the proposed method could operate supervised fine-tuning 
on the pre-trained multilayered network (Fig. \ref{fig:misc}(a)). 
The learning method formed an effective learning representation 
with continuous features, 
and also drastically reorganized the representation in the later layers
(Fig. \ref{fig:misc}(b)).

The proposed learning method was applied to the entire network concurrently 
and not layer by layer. 
Only the correct/incorrect signal was broadcast throughout the network, 
and each local module used just the broadcast signal and 
locally propagated information, 
which is quite suitable for highly distributed parallel computing systems. 
Moreover, the method require no back propagation information, 
decreasing the usage of memory drastically. 
It is critical to process enormously long sequence in deep recurrent networks. 
It could be good 
for application of such a long sequence like natural language. 

No requirement of back propagation 
means more biologically plausible than classical learning methods. 
The error back propagation is sometimes argued its biological unfitness, 
and there are no evidence of its existence in physiological condition. 
The proposed method just utilizes the feedforward signal for the 
association based supervised learning, 
and the advance supervised signal might correspond to 
association by Hebbian rule within the time window of 
spike timing dependent plasticity (STDP) \cite{Markram1997}.

One of the interesting points of the proposed learning method 
is that it seamlessly incorporated 
both reinforcement learning \cite{Sutton98} and competitive learning. 
Reinforcement learning emerges if there is no advance input, 
and the traditional competitive learning emerges if there is no correct/incorrect signal.
This suggests that these learning methods can share the same hardware 
implementation, and 
the learning mode can be selected by the sequence of input and 
correct/incorrect signals. 
Moreover, the timing of the correct/incorrect signal can 
control the associative layer, and it might be useful 
for deeper or recurrent networks.



%

\end{document}